\documentclass[5p,twocolumn]{elsarticle}

\usepackage{amssymb}
\usepackage{amsmath}

\usepackage{algorithm}
\usepackage{algorithmic}
\usepackage{booktabs}
\usepackage{multirow}
\usepackage{subcaption}

\def\Vec#1{{\boldsymbol{#1}}}
\def\Mat#1{{\boldsymbol{#1}}}
\usepackage{footnote}

\journal{Pattern Recognition Letters}

\begin{document}

\begin{frontmatter}



\title{Adaptive Model Ensemble for Continual Learning} 


\author[a]{Yuchuan Mao} 
\author[a]{Zhi Gao}
\author[a]{Xiaomeng Fan}
\author[b,a]{Yuwei Wu}
\author[b,a]{Yunde Jia}
\author[c]{Chenchen Jing}

\affiliation[a]{organization={Beijing Key Laboratory of Intelligent Information Technology}, state={School of Computer Science}, addressline={Beijing Institute of Technology}, country={China}}
\affiliation[b]{organization={Guangdong Laboratory of Machine Perception and Intelligent Computing}, addressline={Shenzhen MSU-BIT University}, country={China}}
\affiliation[c]{organization={Zhejiang University},
            city={Hangzhou},
            country={China}}

\begin{abstract}
Model ensemble is an effective strategy in continual learning, which alleviates catastrophic forgetting by interpolating model parameters, achieving knowledge fusion learned from different tasks. However, existing model ensemble methods usually encounter the knowledge conflict issue at task and layer levels, causing compromised learning performance in both old and new tasks.
To solve this issue, we propose meta-weight-ensembler that adaptively fuses knowledge of different tasks for continual learning. Concretely, we employ a mixing coefficient generator trained via meta-learning to generate appropriate mixing coefficients for model ensemble to address the task-level knowledge conflict. The mixing coefficient is individually generated for each layer to address the layer-level knowledge conflict. In this way, we learn the prior knowledge about adaptively accumulating knowledge of different tasks in a fused model, achieving efficient learning in both old and new tasks. Meta-weight-ensembler can be flexibly combined with existing continual learning methods to boost their ability of alleviating catastrophic forgetting. Experiments on multiple continual learning datasets show that meta-weight-ensembler effectively alleviates catastrophic forgetting and achieves state-of-the-art performance.
\end{abstract}


\begin{keyword}
Continual Learning  \sep Model Ensemble \sep Meta-Learning


\end{keyword}

\end{frontmatter}




\section{Introduction}

\renewcommand{\thefootnote}{}
\footnotetext{\raggedright E-mail addresses: bitmyc@bit.edu.cn (Y. Mao),
gaozhibit@bit.edu.cn (Z. Gao), fanxiaomeng@bit.edu.cn (X. Fan),
wuyuwei@bit.edu.cn (Y. Wu), jiayunde@bit.edu.cn (Y. Jia),
jingchenchen@zju.edu.cn (C. Jing).}

Continual learning aims to imitate the ability of humans to efficiently learn in a dynamic environment with a continuum of tasks, becoming a promising research direction in the computer vision and machine learning communities~\cite{Gao_2024_CVPR,Gao_2023_CVPR,wang2024comprehensive}. In continual learning, model ensemble is an effective strategy to alleviate catastrophic forgetting. It interpolates model parameters learned in different tasks to achieve knowledge fusion during the continual learning process~\cite{Simon_2022_CVPR,Zhang_2022_CVPR,soutifcormerais2023improving}.


\begin{figure}[t]
\centering
\includegraphics[width=0.48\textwidth]{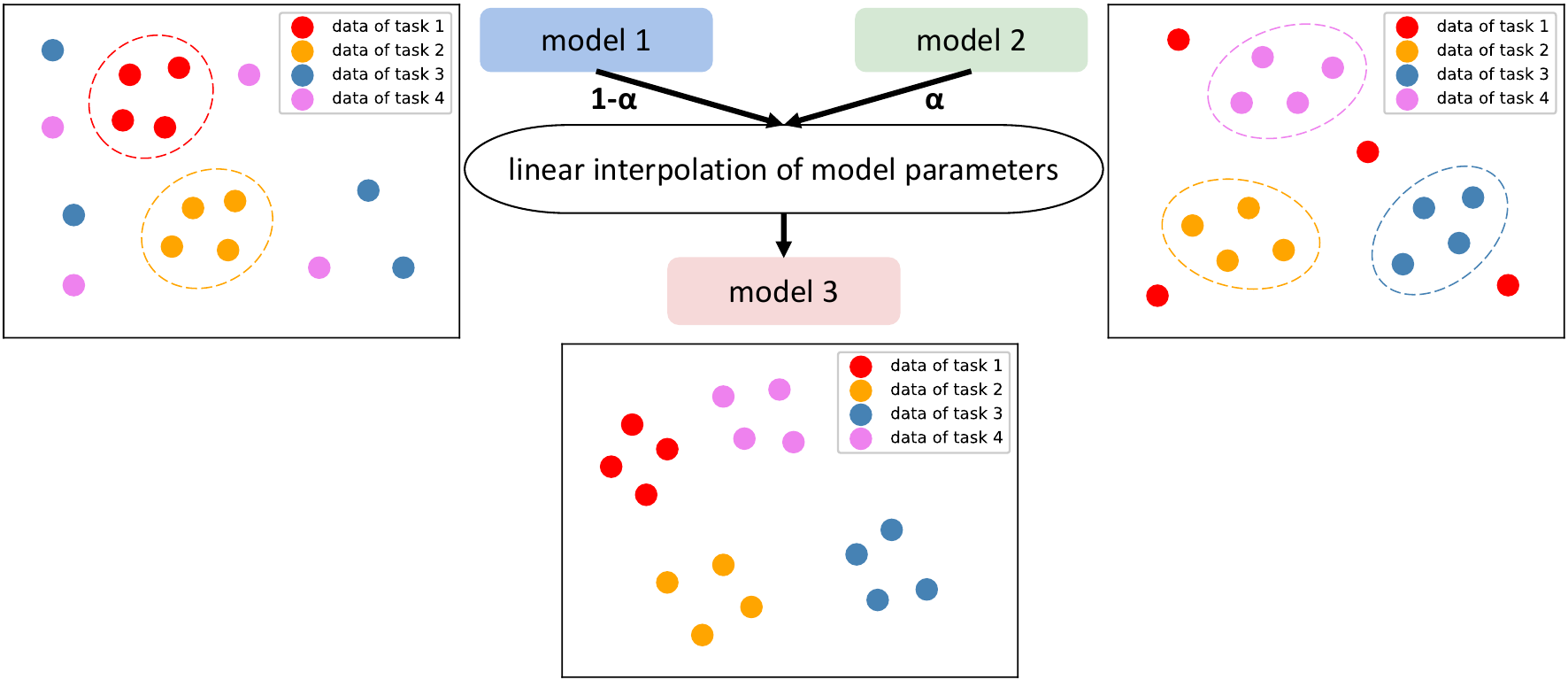}
\vspace{-6mm}
\caption{Illustration of the knowledge conflict issue and our model ensemble method. For model 1 and model 2 which have the same structure, the knowledge of tasks 1-4 is different. Model 1 has knowledge for the classification in tasks 1 and 2, and model 2 has knowledge for the classification in tasks 2, 3 and 4. 
Our goal is to fuse the knowledge in model 1 and model 2 by using the mixing coefficient $\alpha$ to interpolate their parameters. }
\label{fig:image1}
\end{figure}


\begin{figure*}[t]

\centering
\begin{subfigure}{0.22\textwidth}
\includegraphics[width=\linewidth]{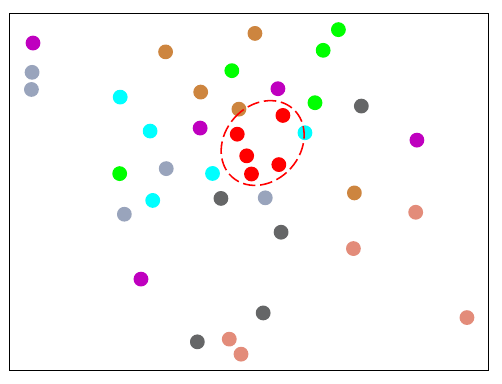}
\caption{Layer 1, Model 1}
\label{fig:subim1}
\end{subfigure}
\begin{subfigure}{0.22\textwidth}
\includegraphics[width=\linewidth]{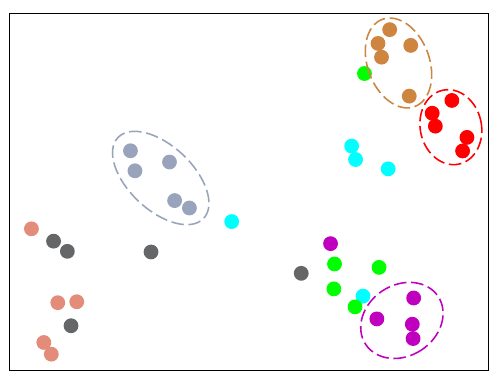}
\caption{Layer 2, Model 1}
\label{fig:subim2}
\end{subfigure}
\begin{subfigure}{0.22\textwidth}
\includegraphics[width=\linewidth]{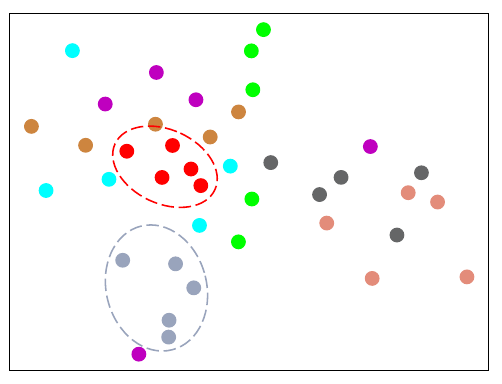}
\caption{Layer 1, Model 2}
\label{fig:subim3}
\end{subfigure}
\begin{subfigure}{0.22\textwidth}
\includegraphics[width=\linewidth]{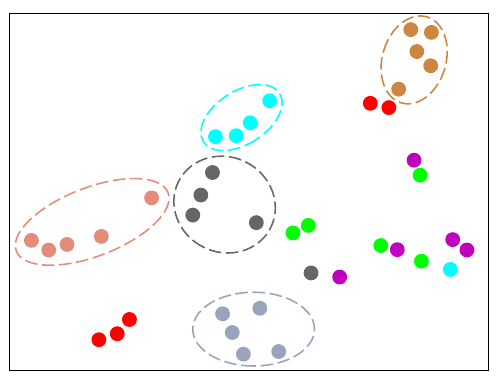}
\caption{Layer 2, Model 2}
\label{fig:subim4}
\end{subfigure}
\vspace{-3mm}
\caption{Comparison between features extracted in different layers of models. 
As shown in (a), (c) and (b), (d), the classification capacity of model 1 and model 2 in various tasks exhibits diversity, suggesting that knowledge of different tasks varies from one another. 
Likewise, we can draw the conclusion that knowledge encoded in different layers of the same model is various when comparing (a), (b) or (c), (d).}
\label{fig:image2}

\end{figure*}


However, there are conflicts between knowledge learned in different tasks. 
For existing model ensemble methods, the task-level and layer-level knowledge conflicts are the issues that ought to be faced when fusing models.
Firstly, a portion of knowledge is unshared between old tasks and new tasks in continual learning. As illustrated in Fig.~\ref{fig:image1}, the classification capacity of models in different tasks exhibits diversity. Model 1 performs better on data of task 1 and task 2, while model 2 works better on data of tasks 2, 3, and 4, suggesting that knowledge of different tasks is various. In this case, directly fusing models trained in different tasks may lead to a conflict of knowledge, which is considered inappropriate and causes compromised learning performance in both old and new tasks.
Secondly, it’s difficult to guarantee that knowledge encoded in different layers of a model plays the same role in continual learning, while it is the prior hypothesis in existing model ensemble methods. As shown in Fig.~\ref{fig:image2}, we compare the classification capacity of different layers in the same model and observe diversity, which suggests that knowledge encoded in different layers of the same model varies from one another. In this case, it is unreasonable to treat models as a whole when alleviating catastrophic forgetting by fusing models in continual learning.
Therefore, it is desirable to build a model ensemble method capable of adaptively fusing knowledge of different tasks at both task and layer levels for continual learning.

To this end, we propose an adaptive model ensemble method: meta-weight-ensembler for continual learning, which alleviates catastrophic forgetting by appropriately fusing knowledge of different tasks. To address the task-level knowledge conflict, we employ a mixing coefficient generator to generate appropriate mixing coefficients for model ensemble with the small amount of data saved in previous tasks. To address the layer-level knowledge conflict, the mixing coefficient is individually generated for each layer. We train the mixing coefficient generator via meta-learning, through which the prior knowledge about generating appropriate mixing coefficients for different tasks and layers is learned from previous tasks.
Moreover, meta-weight-ensembler can be flexibly combined with existing continual learning methods, boosting their ability to alleviate catastrophic forgetting to further improve learning performance.
The main contributions of this paper are listed as follows.

\begin{itemize}
\item We solve the knowledge conflict issue via a novel model ensemble method that adaptively interpolates model parameters of different tasks.
\item We introduce meta-weight-ensembler that utilizes meta-learning to learn to fuse knowledge of different tasks in a data-driven manner.
\end{itemize}


\section{Related work}

Most existing works in continual learning focus on the catastrophic forgetting issue. Catastrophic forgetting is a problem urgent to be solved in continual learning, which is defined as the phenomenon that models forget the knowledge of old tasks after being trained on new tasks.
Catastrophic forgetting in continual learning can be generally divided into three categories: replay-based methods, regularization-based methods and parameter-isolation based methods.

Replay-based methods use data of old tasks saved in a buffer or generated old data to retrain models~\cite{BENKO202465,Goswami_2024_CVPR}. Benkő et al.~\cite{BENKO202465} proposed a simple example selection strategy for replay-based continual learning, better populating the memory of buffers by keeping the least forgettable examples according to forgetting statistics.
Regularization-based methods add regularization terms in loss function to penalize model update~\cite{ZHOU2024137,FRASCAROLI2024119}. Zhou et al.~\cite{ZHOU2024137} proposed to project the gradient of model parameters from old tasks into a designed null space, effectively balancing plasticity and stability in continual learning.
Parameter-isolation based methods allocate different model parameters for different tasks~\cite{miao2022continual,Roy_2024_CVPR,Liang_2024_CVPR}. For example, Miao et al.~\cite{miao2022continual} proposed to decompose the convolutional filters trained in old tasks into atoms that are used to rebuild convolutional filters in the new tasks sharing high similarity with old tasks.

Some continual learning methods use model ensemble as a trick to assist in alleviating catastrophic forgetting, and the core idea of these methods is fusing models trained in different tasks to balance the knowledge of old and new tasks~\cite{Simon_2022_CVPR,Zhang_2022_CVPR,soutifcormerais2023improving}. Simon et al.~\cite{Simon_2022_CVPR} devised an exponential moving average framework for model ensemble in continual learning, which is integrated with learnable projection technique to alleviate catastrophic forgetting.

Existing model ensemble methods directly fuse models trained in different tasks, suffering from the knowledge conflict issue. Different from them, meta-weight-ensembler alleviates catastrophic forgetting by adaptively fusing the knowledge of old tasks and new tasks. In addition, meta-weight-ensembler can be flexibly combined with existing continual learning methods, boosting their ability of alleviating catastrophic forgetting.


\section{Method}

\subsection{Formulation}

In continual learning, the model is required to continually learn from a sequence of tasks $T_1,T_2,\dots,T_t$ in dynamic environment. In learning the $i$-th task $T_i$, only its training data $D_i=\{(\Mat{x_i},\Vec{y_i})\}$ is available, while data of previous tasks is unavailable. 
The goal of continual learning is optimizing parameters $\Theta$ of the model that achieves good performance not only on the current task $T_i$, but also on previous tasks $T_1,T_2,\dots,T_{i-1}$. Continual learning has two popular settings: task-incremental learning and class-incremental learning. In task-incremental learning, task identity is provided during training and test. In contrast, the class-incremental setting is more general, where task identities are not provided at both training and inference. In class-incremental learning, the data of different tasks have no overlap and task identities are not provided at both training and inference. Recently, more and more methods focus on the online class-incremental learning, where training data can only be accessed once.

In this paper, we propose an adaptive model ensemble method: meta-weight-ensembler for continual learning, which adaptively fuses knowledge of different tasks. The illustration of meta-weight-ensembler is in Figure 2. In the $i$-th task, we have two models. One is the model trained in the current task $T_{i}$ and the other is the model trained in the last task $T_{i-1}$. The model trained in $T_{i}$ is represented as $f_{\hat{\Theta}_i}$, which accumulates the knowledge of $T_{i}$. The model trained in $T_{i-1}$ is represented as $f_{\Theta_{i-1}}$, which accumulates the knowledge of old tasks $T_1,T_2,\dots,T_{i-1}$. Our goal is appropriately fusing the knowledge accumulated in $f_{\hat{\Theta}_i}$ and $f_{\Theta_{i-1}}$ to obtain the model applicable for all existing tasks $T_1,T_2,\dots,T_i$, the process of which can be formulated as 

\begin{equation}
    f_{\Theta_i}= M_{\Phi} ( f_{\hat{\Theta}_i}, f_{\Theta_{i-1}} ),
\end{equation}
where $M_{\Phi}$ represents meta-weight-ensembler, and $\Phi$ represents its parameters. To be specific, we employ a mixing coefficient generator to adaptively generate the independent mixing coefficient for parameters of each layer in the model. Then, we fuse the knowledge accumulated in $f_{\hat{\Theta}_i}$ and $f_{\Theta_{i-1}}$ by linearly interpolating their parameters in a layer-wise manner. We train the mixing coefficient generator via meta-learning, seeking the way to explore knowledge on all existing tasks, and generate appropriate mixing coefficients for all tasks.


\begin{figure}[t]
\centering
\includegraphics[width=0.47\textwidth]{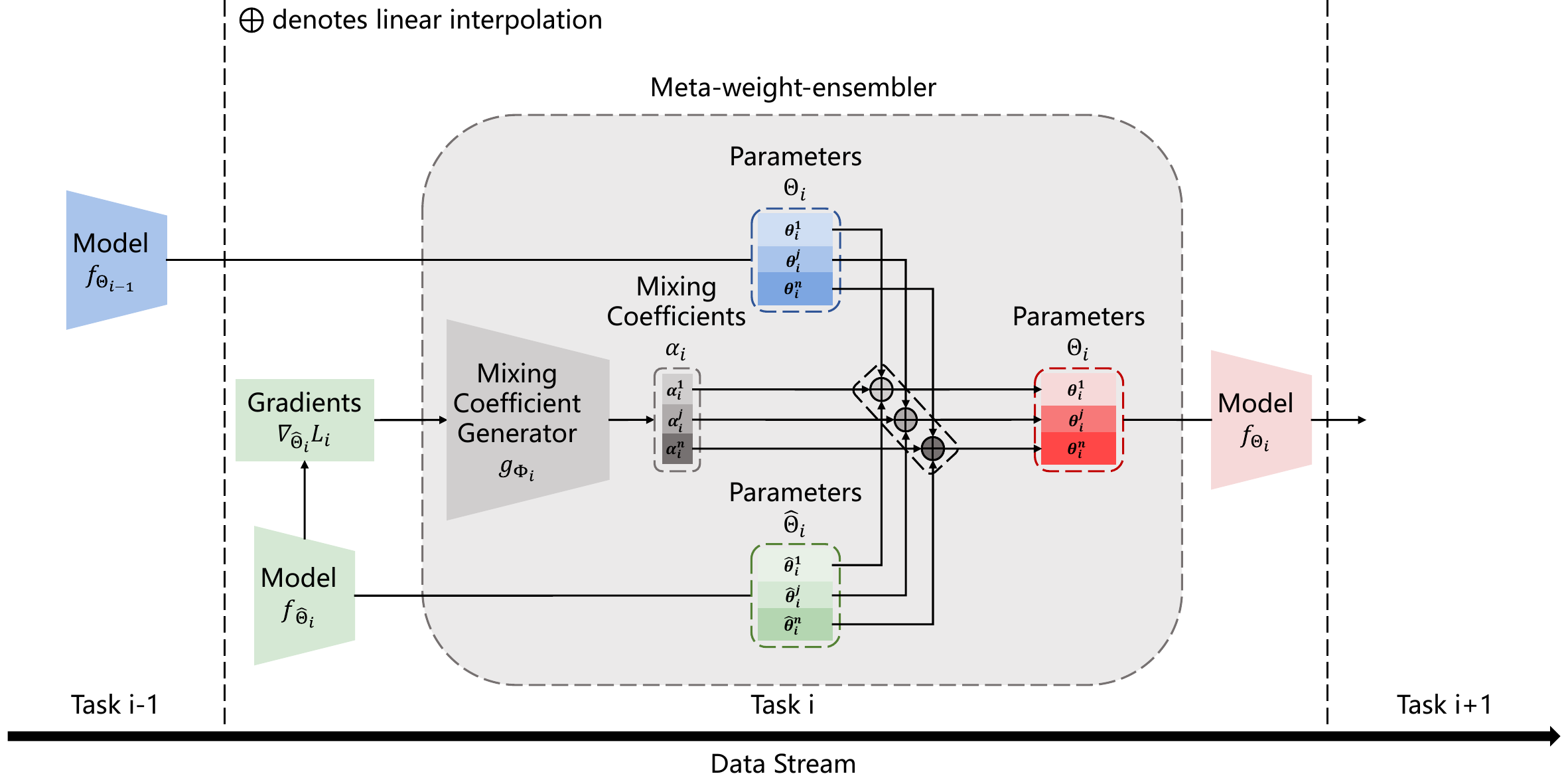}
\vspace{-3mm}
\caption{Formulation of Meta-weight-ensembler}
\label{fig:image3}
\end{figure}


\subsection{Model Ensemble in Layer-wise Manner}

Meta-weight-ensembler fuses knowledge of different tasks by linearly interpolating parameters of the models trained in different tasks. For a model $f_{\Theta}$, we represent parameters of the $j$-th layer as $\Vec{\theta^j}$. In the $i$-th task, parameters of the model trained in current task $T_i$ is represented as $\hat{\Theta}_i=\{\Vec{\hat{\theta}_i^1},\Vec{\hat{\theta}_i^2},\dots,\Vec{\hat{\theta}_i^n}\}$, and parameters of the model trained in the last task $T_{i-1}$ is represented as $\Theta_{i-1}=\{\Vec{\theta_{i-1}^1},\Vec{\theta_{i-1}^2},\dots,\Vec{\theta_{i-1}^n}\}$. We use $\alpha_i=\{\alpha_i^1,\alpha_i^2,\dots,\alpha_i^n\}$ to represent the independent mixing coefficients for parameters of all layers in the model.

Given $\hat{\Theta}_i$ and $\Theta_{i-1}$, meta-weight-ensembler uses $\alpha_i$ to fuse the model trained in current task $T_i$ and the model trained in the last task $T_{i-1}$ by linearly interpolating their parameters in a layer-wise manner. The process of model fusion is formulated as

\begin{equation}
    \Vec{\theta_i^j}=\alpha_i^j\cdot\Vec{\hat{\theta}_i^j}+(1-\alpha_i^j)\cdot\Vec{\theta_{i-1}^j}.\qquad(j\in[1, n])
\end{equation}

The parameters $\Theta_i=\{\Vec{\theta_i^1},\Vec{\theta_i^2},\dots,\Vec{\theta_i^n}\}$ of the fused model are used as initialization parameters of the model in the next task $T_{i+1}$. Since knowledge encoded in different layers of the same model is various, meta-weight-ensembler achieves more appropriate fusion of knowledge compared with model ensemble methods treating models as a whole.
Taking the diverse existing continual learning methods into consideration, weight-space ensemble is a natural choice for a general knowledge fusion mechanism to alleviate catastrophic forgetting as it is model-agnostic and ensembles without extra computational cost.

\subsection{Mixing Coefficient Generator}

Meta-weight-ensembler employs a mixing coefficient generator to adaptively generate the independent mixing coefficient for parameters of each layer in the model. The mixing coefficient generator is a multilayer perceptron consisting of two linear layers, the dimension of whose output is equal to the number of layers in the model. 
To make mixing coefficients specific to the current task and layer, we take gradients as the input of the mixing coefficient generator. Gradients, used for the optimization of model parameters, hold task-specific optimization information~\cite{Baik_2020_CVPR}. Gradients represent diversity in the current model and old models, reflecting the disparity between knowledge of different tasks. Thus, we compute the mean gradients of different layers after the model training in each task to generate mixing coefficients specific to the current task and layers in the current model.

In the $i$-th task, we have the data $D_i$ and the loss function $L_i$ of current task $T_i$, which can be used to compute gradients $\nabla_{\hat{\Theta}_i} L_i$ of the model $f_{\hat{\Theta}_i}$. We represent the mixing coefficient generator as $g_{\Phi_i}$, where $\Phi_i$ represents its parameters.
Given $D_i$ and $L_i$, $g_{\Phi_i}$ takes $\nabla_{\hat{\Theta}_i} L_i$ as input to generate independent mixing coefficients $\alpha_i$ for parameters of all layers in the model, which can be formulated as follows,

\begin{equation}
    \alpha_i=g_{\Phi_i}(\nabla_{\hat{\Theta}_i} L_i).
\end{equation}


\begin{algorithm}[t]
\caption{Meta-weight-ensembler}
\label{alg:algorithm}
\textbf{Input}: Data $D_1,D_2,\dots,D_t$\\
\textbf{Output}: Base Learner Parameters $\Theta_t$\\
\textbf{Initialize}: $\hat{\Theta}_{i-1}=\Theta_{0}$, $\Phi_{i-1}=\Phi_{0}$,\\$\hat{D_i}=\{\}$
\begin{algorithmic}[1] 
\FOR {$i=1,2,\dots,t$}
\STATE $\hat{\Theta}_{i}\leftarrow\Theta_{i-1}$, $\Phi_i\leftarrow\Phi_{i-1}$
\STATE Train $f_{\hat{\Theta}_i}$ on $D_i$
\IF{$i>1$}
\FOR {$m=1,2,\dots,iteration\_ num$}
\STATE Generate $\alpha_i\leftarrow g_{\Phi_i}(\nabla_{\hat{\Theta}_i} L_i)$
\FOR {$j=1,2,\dots,n$}
\STATE Obtain $\Vec{\theta_i^j}\leftarrow\alpha_i^j\cdot\Vec{\hat{\theta}_i^j}+(1-\alpha_i^j)\cdot\Vec{\theta_{i-1}^j}$
\ENDFOR
\STATE Compute $\nabla_{\Phi_i}\hat{L}_i$ using $\hat{D}_i$
\STATE Use $\nabla_{\Phi_i}\hat{L}_i$ to update $\Phi_i$ in $g_{\Phi_i}$
\ENDFOR
\ENDIF
\STATE Save part of $D_i$ into $\hat{D_i}$
\STATE Obtain $\Theta_i=\{\theta_i^1,\theta_i^2,\dots,\theta_i^n\}$
\ENDFOR
\STATE \textbf{return} $\Theta_i$
\end{algorithmic}
\end{algorithm}


\subsection{Training}

We introduce meta-learning to train the mixing coefficient generator, where a bi-level optimization framework is employed. In the $i$-th task, we take the data $D_i=\{(\Mat{x_i},\Vec{y_i}\}$ of current task $T_i$ as the training data, and a small part of the data saved in all existing tasks $T_1,T_2,\dots,T_i$ is used as the validation data $\hat{D}_i=\{(\Mat{\hat{x}_i},\Vec{\hat{y}_i})\}$. 
$L_i$ represents loss functions on data $D_i$, and $\hat{L}_i$ represents loss functions corresponding to the data in $\hat{D}_i$. 
Before training the mixing coefficient generator, we train  $f_{\hat{\Theta}_i}$ to converge by minimizing $L_i$.
Then, we update the mixing coefficient generator in a bi-level optimization manner. In the inner-loop, given the loss function $L_i$ of the current task, the model $f_{\hat{\Theta}_i}$ trained in the current task, and the model $f_{\Theta_{i-1}}$ trained in the last task, the mixing coefficient generator $g_{\Phi_i}$ takes gradients $\nabla_{\hat{\Theta}_i} L_i$ of $f_{\hat{\Theta}_i}$ for $D_i$ as input to generate the independent mixing coefficient $\alpha_i$ for parameters of each layer in the model, then obtains parameters $\Theta_{i}$ of the fused model by using $\alpha_i$ to linearly interpolate parameters $\hat{\Theta}_i$ of $f_{\hat{\Theta}_i}$ and parameters $\Theta_{i-1}$ of $f_{\Theta_{i-1}}$. In the outer-loop, we update $g_{\Phi_i}$ on $\hat{D}_i$ for a specific number $M$ of times by minimizing $\hat{L}_i$ and obtain the updated mixing coefficient generator $g_{\Phi_i^*}$,
\begin{equation}
\begin{aligned}
&\Phi_i^*=\arg \min_{\Phi_i}E_{(\Mat{\hat{x}_i},\Vec{{y}_i})\in \hat{D}_i}\hat{L}_i(\Vec{\hat{y}_i},f_{\Theta_i}(\Mat{\hat{x}_i})).\\
&s.t.\qquad\Theta_i=g_{\Phi_i}(\nabla_{\hat{\Theta}_i} L_i)\cdot\hat{\Theta}_i+(1-g_{\Phi_i}(\nabla_{\hat{\Theta}_i} L_i))\cdot\Theta_{i-1}
\end{aligned}
\end{equation}

By learning the prior knowledge about generating mixing coefficients, meta-weight-ensembler alleviates catastrophic forgetting at both task and layer levels in an adaptive manner to appropriately fuses knowledge of different tasks, improving performance of the model on both old and new tasks. 
Since the method to obtain $f_{\hat{\Theta}_i}$ in the current task can be replaced by other continual learning methods, meta-weight-ensembler can be taken as a general knowledge fusion mechanism which is flexibly combined with existing continual learning methods, further improving their learning performance by boosting their ability of alleviating catastrophic forgetting. Algorithm 1 summarizes training details of meta-weight-ensembler.


\section{Experiments}

\subsection{Experiment Setup}


\textbf{Datasets.} We conducted experiments on three continual learning datasets: Split CIFAR-10, Split CIFAR-100 and Split MiniImageNet. (1) Split CIFAR-10 splits CIFAR-10~\cite{krizhevsky2009learning} into 5 tasks, each with 2 classes. (2) Split CIFAR-100 splits CIFAR-100~\cite{krizhevsky2009learning} into 10 tasks, each with 10 classes. (3) Split MiniImageNet splits MiniImageNet~\cite{NIPS2016_90e13578} into 10 tasks, each with 10 classes. The MiniImageNet dataset is a subset of ImageNet~\cite{5206848} with 100 classes. For the three used datasets, we follow standard settings in continual learning to split training and test sets.


\begin{table*}[t] 
\centering
\caption{Results for Task-incremental Learning} 
\vspace{-3mm}
\begin{tabular}{*{7}{c}}
  \toprule
  \multirow{2}*{Method} & \multicolumn{2}{c}{Split CIFAR-10} & \multicolumn{2}{c}{Split CIFAR-100} & \multicolumn{2}{c}{Split MiniImageNet} \\
  \cmidrule(lr){2-3}\cmidrule(lr){4-5}\cmidrule(lr){6-7}
  & ACC (\%) & BWT (\%) & ACC (\%) & BWT (\%) & ACC (\%) & BWT (\%) \\
  \midrule
  ER~\cite{NEURIPS2019_fa7cdfad} & 90.60 & -7.74 & 66.82 & -22.73 & 28.97 & -28.40 \\
  GMED~\cite{NEURIPS2021_f45a1078} & 89.72 & -8.75 & 68.82 & -20.53 & 30.47 & -26.02 \\
  MetaSP~\cite{NEURIPS2022_ad2fa437} & 91.41 & -7.36 & 70.81 & -19.74 & 34.36 & -21.70 \\
  CLS-ER~\cite{arani2022learning} & 92.86 & -5.00 & 65.55 & -16.2 & 40.21 & \textbf{-5.03} \\
  InfluenceCL~\cite{Sun_2023_CVPR} & 92.53 & -5.46 & 72.53 & -17.22 & 36.46 & -19.48 \\
  \textbf{InfluenceCL+Ours} & \textbf{93.04} & \textbf{-2.33} & \textbf{76.04} & \textbf{-9.12} & \textbf{44.44} & -13.40 \\
  \midrule
  iCaRL~\cite{Rebuffi_2017_CVPR} & 90.27 & -4.29 & 84.40 & -3.72 & 63.62 & \textbf{-12.23} \\
  FDR~\cite{benjamin2018measuring} & 91.42 & -7.03 & 74.66 & -16.63 & 63.20 & -27.69 \\
  LUCIR~\cite{Hou_2019_CVPR} & 94.30 & -2.83 & 84.41 & \textbf{-2.42} & 68.14 & -15.08 \\
  BFP~\cite{Gu_2023_CVPR} & 94.66 & -4.15 & 82.31 & -11.57 & 62.00 & -18.97 \\
  \textbf{BFP+Ours} & \textbf{95.14} & \textbf{-2.71} & \textbf{84.44} & -8.73 & \textbf{68.42} & -17.50 \\
  \midrule
  META~\cite{Xue_2022_CVPR} & 95.83 & -3.53 & 45.12 & -51.29 & 60.94 & -35.14 \\
  \textbf{MEAT+Ours} & \textbf{98.20} & \textbf{-1.23} & \textbf{54.61} & \textbf{-41.39} & \textbf{68.81} & \textbf{-27.26} \\
  \bottomrule
\end{tabular}
\label{taskincrementalLearning}
\end{table*}


\begin{table*}[t] 
\centering
\caption{Results for Class-incremental Learning} 
\vspace{-3mm}
\begin{tabular}{*{7}{c}}
  \toprule
  \multirow{2}*{Method} & \multicolumn{2}{c}{Split CIFAR-10} & \multicolumn{2}{c}{Split CIFAR-100} & \multicolumn{2}{c}{Split MiniImageNet} \\
  \cmidrule(lr){2-3}\cmidrule(lr){4-5}\cmidrule(lr){6-7}
  & ACC (\%) & BWT (\%) & ACC (\%) & BWT (\%) & ACC (\%) & BWT (\%) \\
  \midrule
  ER~\cite{NEURIPS2019_fa7cdfad} & 40.45 & -70.36 & 13.75 & -81.64 & 11.00 & -50.84 \\
  GMED~\cite{NEURIPS2021_f45a1078} & 43.68 & -66.21 & 14.56 & -80.68 & 11.03 & -50.23 \\
  MetaSP~\cite{NEURIPS2022_ad2fa437} & 50.10 & -58.39 & 19.28 & -76.13 & 12.74 & -48.84 \\
  CLS-ER~\cite{arani2022learning} & 62.94 & -41.90 & 11.50 & \textbf{-17.82} & 9.29 & \textbf{-3.00} \\
  InfluenceCL~\cite{Sun_2023_CVPR} & 53.07 & -54.44 & 21.15 & -73.24 & 13.63 & -47.94 \\
  \textbf{InfluenceCL+Ours} & \textbf{64.78} & \textbf{-10.01} & \textbf{27.50} & -56.27 & \textbf{14.97} & -21.02 \\
  \midrule
  iCaRL~\cite{Rebuffi_2017_CVPR} & 63.58 & -27.75 & 46.66 & -30.13 & 29.46 & -28.51 \\
  FDR~\cite{benjamin2018measuring} & 31.24 & -76.08 & 22.64 & -73.71 & 26.76 & -75.20 \\
  LUCIR~\cite{Hou_2019_CVPR} & 58.53 & -46.36 & 35.14 & -53.24 & 41.46 & -34.84 \\
  BFP~\cite{Gu_2023_CVPR} & 78.71 & -28.28 & 47.45 & -29.85 & 38.34 & -35.90 \\
  \textbf{BFP+Ours} & \textbf{81.33} & \textbf{-20.85} & \textbf{61.19} & \textbf{-26.91} & \textbf{43.59} & \textbf{-25.02} \\
  \midrule
  MEAT~\cite{Xue_2022_CVPR} & 19.88 & -99.39 & 9.68 & -95.69 & 9.45 & -95.4 \\
  \textbf{MEAT+Ours} & \textbf{47.95} & \textbf{-50.49} & \textbf{21.97} & \textbf{-44.44} & \textbf{33.45} & \textbf{-40.20} \\
  \bottomrule
\end{tabular}
\label{classincrementalLearning}
\end{table*}


\begin{table*}[t] 
\centering
\caption{Results for Online Class-incremental Learning} 
\vspace{-3mm}
\begin{tabular}{*{7}{c}}
  \toprule
  \multirow{2}*{Method} & \multicolumn{2}{c}{Split CIFAR-10} & \multicolumn{2}{c}{Split CIFAR-100} & \multicolumn{2}{c}{Split MiniImageNet} \\
  \cmidrule(lr){2-3}\cmidrule(lr){4-5}\cmidrule(lr){6-7}
  & ACC (\%) & BWT (\%) & ACC (\%) & BWT (\%) & ACC (\%) & BWT (\%) \\
  \midrule
  ER~\cite{NEURIPS2019_fa7cdfad} & 41.7 & - & 17.6 & - & 13.4 & - \\
  GMED~\cite{NEURIPS2021_f45a1078} & 43.6 & - & 18.8 & - & 15.3 & - \\
  ER-WA~\cite{Zhao_2020_CVPR} & 42.5 & - & 21.7 & - & 17.1 & - \\
  DER~\cite{NEURIPS2020_b704ea2c} & 45.3 & - & 17.2 & - & 14.8 & - \\
  SS-IL~\cite{Ahn_2021_ICCV} & 42.2 & - & 21.9 & - & 19.7 & - \\
  SCR~\cite{Mai_2021_CVPR} & 45.4 & - & 16.2 & - & 14.7 & - \\
  ER-DVC~\cite{Gu_2022_CVPR} & 45.4 & - & 19.7 & - & 15.4 & - \\
  OCM~\cite{pmlr-v162-guo22g} & 49.9 & - & 20.6 & - & 13.6 & - \\
  ER-ACE~\cite{caccia2022new} & 49.7 & - & 23.1 & - & 20.3 & - \\
  CBA~\cite{cba} & 32.57 & -24.97 & 22.25 & -10.57 & 14.29 & -14.84 \\
  ASER~\cite{Shim_Mai_Jeong_Sanner_Kim_Jang_2021} & 37.22 & -51.89 & 21.73 & -30.91 & 18.89 & -20.95 \\
  \textbf{ASER+Ours} & \textbf{42.97} & \textbf{-23.34} & \textbf{28.69} & \textbf{-5.86} & \textbf{22.05} & \textbf{-5.40} \\
  PCR~\cite{Lin_2023_CVPR} & 50.70 & -23.38 & 23.40 & -20.49 & 25.10 & -14.49 \\
  \textbf{PCR+Ours} & \textbf{55.40} & \textbf{-13.65} & \textbf{27.30} & \textbf{-7.75} & \textbf{25.30} & \textbf{-7.46} \\
  \bottomrule
\end{tabular}
\label{onlineclassincrementalLearning}
\end{table*}



\textbf{Baselines.} For the evaluation of capability to alleviate catastrophic forgetting in three continual learning settings, we combine meta-weight-ensembler with five state-of-the-art continual learning methods from three different categories of continual learning methods. 
We select three methods for task-incremental learning (TIL) and class-incremental learning (CIL), where InfluenceCL~\cite{Sun_2023_CVPR} belongs to replay-based methods, BFP~\cite{Gu_2023_CVPR} is an advanced regularization-based method, and MEAT~\cite{Xue_2022_CVPR} belongs to architecture-based methods.
Since replay-based methods are the main solutions of online class-incremental learning (OCIL), we choose ASER~\cite{Shim_Mai_Jeong_Sanner_Kim_Jang_2021} and PCR~\cite{Lin_2023_CVPR} for OCIL.


\textbf{Metrics.} We employ two evaluation metrics proposed by Lopez-Paz et al.~\cite{NIPS2017_f8752278} to evaluate the ability of meta-weight-ensembler to alleviate catastrophic forgetting: Average Accuracy (ACC) and Backward Transfer (BWT).
(1) $ACC=\frac{1}{T}\Sigma^T_{i=1}R_{T,i}$ is the average test accuracy of the model on each task after trained in all tasks.
(2) $BWT=\frac{1}{T}\Sigma^{T-1}_{i=1}R_{T,i}-R_{i,i}$ is a negative value representing the average forgetting of all previous tasks, the smaller absolute value of which indicates the better learning performance.
$R_{i,j}$ represents the test accuracy of the model on task $t_j$ after observing the last sample from task $t_i$, and $T$ represents the number of tasks appear in the sequence.


\textbf{Backbones.} We adopt ViT~\cite{dosovitskiy2021an} for MEAT in all three continual learning settings. For InfluenceCL, BFP, ASER and PCR, we adopt ResNet-18~\cite{He_2016_CVPR} for TIL and CIL, and adopt Reduced ResNet-18~\cite{Lin_2023_CVPR} for OCIL.


\subsection{Main Results}

We illustrate results of TIL and CIL in Table~\ref{taskincrementalLearning} and ~\ref{classincrementalLearning}. Overall, meta-weight-ensembler significantly improves the learning performance of baseline methods when combined with them and achieves the best learning performance on three datasets in terms of two metrics. These experimental results show the state-of-the-art capability of meta-weight-ensembler in alleviating catastrophic forgetting.
For example, in CIL of Table~\ref{classincrementalLearning}, meta-weight-ensembler improves the learning performance of InfluenceCL on different datasets by $16.97\%$-$44.43\%$ in terms of BWT.



We illustrate experimental results of OCIL in Table~\ref{onlineclassincrementalLearning}, which also show significant improvements in baseline methods. From the comprehensive results shown in the three settings, we conclude that meta-weight-ensembler can adaptively fuse knowledge in continual learning.



\begin{figure*}[t]
\centering
\begin{subfigure}{0.22\textwidth}
\includegraphics[width=\linewidth]{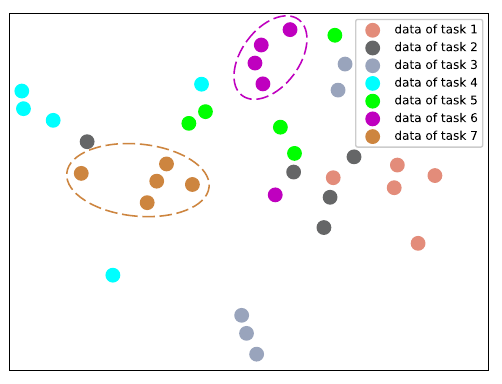} 
\caption{Fused Model of Task i-1} 
\label{fig:subim5}
\end{subfigure}
\begin{subfigure}{0.22\textwidth}
\includegraphics[width=\linewidth]{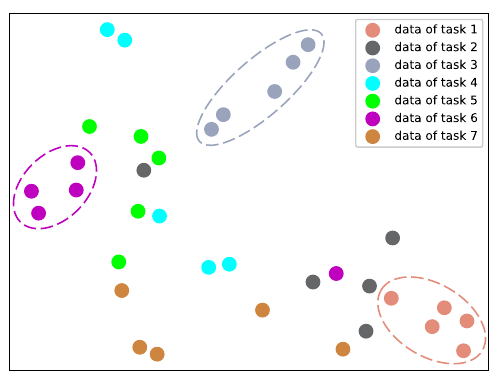}
\caption{Trained Model of Task i} 
\label{fig:subim6}
\end{subfigure}
\begin{subfigure}{0.22\textwidth}
\includegraphics[width=\linewidth]{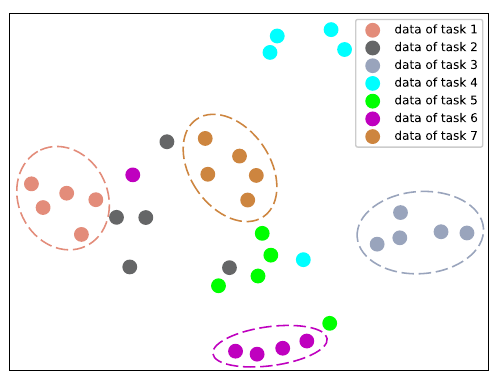} 
\caption{Fused Model of Task i} 
\label{fig:subim7}
\end{subfigure}
\vspace{-3mm}
\caption{Comparison between features extracted by the fused model in the last task, the model trained in the current task and the fused model in the current task. The models are trained by `InfluenceCL+Ours'. As shown in figures, the fused model in the current task has the classification capacity owned by the fused model in the last task and the model trained in the current task, suggesting that our method appropriately fuses knowledge in different models.}
\label{fig:image4}
\end{figure*}


\subsection{Ablation}

We conduct ablation experiments using Split CIFAR-100 dataset in TIL and CIL to show the effectiveness of our meta-weight-ensembler. 
Firstly, we set the mixing coefficient as $0.5$ for all layers, denoted by `E'. Then, we generate one mixing coefficient for all layers, denoted by `E+ML'. Finally, we independent generate the mixing coefficient for each layer, denoted by `E+ML+LW'. 
From the results shown in Table~\ref{ablationExperiments}, we can observe that `E+ML' outperforms `E' in most cases, which indicates that the bi-level optimization framework improves the capability of weight-space ensemble to effectively alleviating catastrophic forgetting in continual learning. We can also observe that `E+ML+LW' achieves the best learning performance in terms of ACC. The reason is that it is unreasonable to treat models as a whole when fusing knowledge in continual learning. In contrast, meta-weight-ensembler fuses models in a layer-wise manner. 
We find that `E+ML+LW' has lower BWT than 'E+ML' in class-incremental learning. The reason is that layer-wise ensemble manner is the reasonable way to fuse knowledge in continual learning, which provides a higher historical highest accuracy, i.e. $R_{i,i}$ in the definition formula, for each task as shown in Table~\ref{historicalHighestAccuracy}.


\begin{table}[h] 
\centering
\caption{Results of Ablation Experiments}
\vspace{-3mm}
\scalebox{0.95}{
\begin{tabular}{*{5}{c}}
  \toprule
  \multirow{2}*{Method} & \multicolumn{2}{c}{Task-incremental} & \multicolumn{2}{c}{Class-incremental} \\
  \cmidrule(lr){2-3}\cmidrule(lr){4-5}
  & ACC & BWT & ACC & BWT \\
  \midrule
  E & 90.42 & -5.41 & 52.51 & -8.56 \\
  E+ML & 92.10 & -3.70 & 60.34 & \textbf{-7.66} \\
  E+ML+LW & \textbf{93.88} & \textbf{-2.46} & \textbf{64.65} & -20.78 \\
  \bottomrule
\end{tabular}
}
\label{ablationExperiments}
\end{table}


\begin{table}[h] 
\centering
\caption{Historical Highest Accuracy in Class-incremental Learning}
\vspace{-3mm}
\scalebox{0.95}{
\begin{tabular}{*{6}{c}}
  \toprule
  Method & Task 2 & Task 3 & Task 4 & Task 5 \\
  \midrule
  E & 51.20 & 34.95 & 57.45 & 54.25 \\
  E+ML & 44.85 & 45.25 & 80.30 & 63.40 \\
  E+ML+LW & \textbf{75.80} & \textbf{90.50} & \textbf{94.70} & \textbf{91.00} \\
  \bottomrule
\end{tabular}
}
\label{historicalHighestAccuracy}
\end{table}


\begin{figure}[h]
\centering
\begin{subfigure}{0.22\textwidth}
\includegraphics[width=\linewidth]{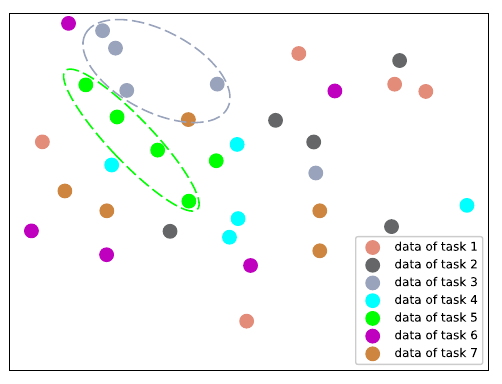}
\caption{the Penultimate Block}
\label{fig:subim8}
\end{subfigure}
\begin{subfigure}{0.22\textwidth}
\includegraphics[width=\linewidth]{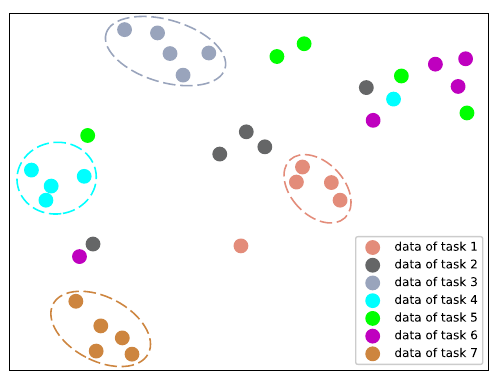}
\caption{the Last Block}
\label{fig:subim9}
\end{subfigure}
\vspace{-3mm}
\caption{Comparison between features extracted by the penultimate and the last block of Resnet-18 trained by `InfluenceCL+Ours'. As shown in figures, features extracted by different layers have good classification performance on different tasks, and the features extracted by deep layers exhibit good classification performance on more tasks, suggesting that knowledge encoded in various layers is different.}
\label{fig:image5}
\end{figure}


\subsection{Visualization}

Based on the models trained by ‘InfluenceCL+Ours’, We plot the features extracted by the fused model in the last task, the model trained in the current task and the fused model in the current task respectively as shown in Fig.~\ref{fig:image4}. A small dotted circle means that the model performs well on the classification task with corresponding colour.
As shown in Fig.~\ref{fig:image4}, the fused model in the current task has the classification capacity owned by the fused model in the last task and the model trained in the current task. Results show that the model fused via meta-weight-ensembler appropriately accumulates the knowledge of the current task and the last task, alleviating catastrophic forgetting at both task and layer levels.

We also visualize the features extracted by the penultimate and the last block of Resnet-18 trained by `InfluenceCL+Ours'.
Fig.~\ref{fig:image5} shows that features extracted by different layers have good classification performance on different tasks, and the features extracted by deep layers exhibit good classification performance on more tasks, reflecting knowledge encoded in various layers is different.


\section{Conclusions}

In this paper, we have presented an adaptive model ensemble method: meta-weight-ensembler for continual learning, which alleviates catastrophic forgetting by appropriately fusing knowledge of different tasks. 
The designed layer-wise manner for model ensemble is capable of achieving adaptive knowledge fusion, alleviating the knowledge conflict in different tasks and different layers. 
The employed mixing coefficient generator can explore the variation of data, benefiting to produce suitable mixing coefficients.
In addition, the mixing coefficient generator is trained in the meta-learning framework, which accumulates the prior information about generating mixing coefficients from previously learned tasks, and then applies the accumulated information to new tasks for suitable mixing coefficients.
Experimental results on multiple continual learning datasets show the effectiveness of our method.



\bibliographystyle{elsarticle-num-names}
\bibliography{main}
\end{document}